\documentclass[sigconf]{acmart}
\usepackage{graphicx}
\usepackage{amsmath}
\usepackage{booktabs}
\usepackage{multirow}
\usepackage{color}
\usepackage{bbding}
\usepackage{microtype}
\usepackage{bm}
\usepackage[american]{babel}
\usepackage{wasysym}
\usepackage{indentfirst}
\usepackage{appendix}
\AtBeginDocument{%
 \providecommand\BibTeX{{%
 \normalfont B\kern-0.5em{\scshape i\kern-0.25em b}\kern-0.8em\TeX}}}

\acmSubmissionID{426}

\copyrightyear{2023}
\acmYear{2023}
\setcopyright{acmlicensed}
\acmConference[MM '23] {Proceedings of the 31st ACM International Conference on Multimedia}{October 29--November 3, 2023}{Ottawa, ON, Canada.}
\acmBooktitle{Proceedings of the 31st ACM International Conference on Multimedia (MM '23), October 29--November 3, 2023, Ottawa, ON, Canada}
\acmPrice{15.00}
\acmISBN{979-8-4007-0108-5/23/10}
\acmDOI{10.1145/3581783.3611793}

\begin{document}\sloppy

\title{DAOT: Domain-Agnostically Aligned Optimal Transport for Domain-Adaptive Crowd Counting}

\author{Huilin Zhu}
\affiliation{%
	\institution{School of Computer Science and Artificial Intelligence,\\Wuhan University of Technology}
	\city{}
	\state{}
	\country{}
}
\email{jsj_zhl@whut.edu.cn}

\author{Jingling Yuan}
\affiliation{%
	\institution{School of Computer Science and Artificial Intelligence,\\Wuhan University of Technology}
	\city{}
	\state{}
	\country{}
}
\email{yjl@whut.edu.cn}

\author{Xian Zhong}
\authornote{Corresponding authors.}
\affiliation{%
	\institution{School of Computer Science and Artificial Intelligence,\\Wuhan University of Technology}
	\city{}
	\state{}
	\country{}
}
\email{zhongx@whut.edu.cn}

\author{Zhengwei Yang}
\affiliation{%
	\institution{National Engineering Research Center for Multimedia Software, \\ School of Computer Science, \\ Wuhan University}
	\city{}
	\state{}
	\country{}
}
\email{yzw_aim@whu.edu.cn}

\author{Zheng Wang}
\affiliation{%
	\institution{National Engineering Research Center for Multimedia Software, \\ School of Computer Science, \\ Wuhan University}
	\city{}
	\state{}
	\country{}
}
\email{wangzwhu@whu.edu.cn}

\author{Shengfeng He}
\authornotemark[1]
\affiliation{%
	\institution{School of Computing and \\Information Systems,\\Singapore Management University}
	\city{}
	\state{}
	\country{}
}
\email{shengfenghe@smu.edu.sg}
\renewcommand{\shortauthors}{Huilin Zhu, Jingling Yuan, Xian Zhong, Zhengwei Yang, Zheng Wang, \&~Shengfeng He}

\newcommand\blfootnote[1]{%
\begingroup 
\hspace{-0.6em}
\renewcommand\thefootnote{}\footnote{#1}%
\addtocounter{footnote}{-1}%
\endgroup 
}
\begin{abstract}
Domain adaptation is commonly employed in crowd counting to bridge the domain gaps between different datasets. However, existing domain adaptation methods tend to focus on inter-dataset differences while overlooking the intra-differences within the same dataset, leading to additional learning ambiguities. These domain-agnostic factors, \textit{e.g.}, density, surveillance perspective, and scale, can cause significant in-domain variations, and the misalignment of these factors across domains can lead to a drop in performance in cross-domain crowd counting. To address this issue, we propose a Domain-agnostically Aligned Optimal Transport (DAOT) strategy that aligns domain-agnostic factors between domains. 
The DAOT consists of three steps. First, individual-level differences in domain-agnostic factors are measured using structural similarity (SSIM). Second, the optimal transfer (OT) strategy is employed to smooth out these differences and find the optimal domain-to-domain misalignment, with outlier individuals removed via a virtual ``dustbin'' column. Third, knowledge is transferred based on the aligned domain-agnostic factors, and the model is retrained for domain adaptation to bridge the gap across domains.
We conduct extensive experiments on five standard crowd-counting benchmarks and demonstrate that the proposed method has strong generalizability across diverse datasets. Our code will be available at: \url{https://github.com/HopooLinZ/DAOT/}.
\end{abstract}


\maketitle

\section{Introduction}

\begin{figure}[!t]
	\centering
	\includegraphics[width = 0.9\columnwidth]{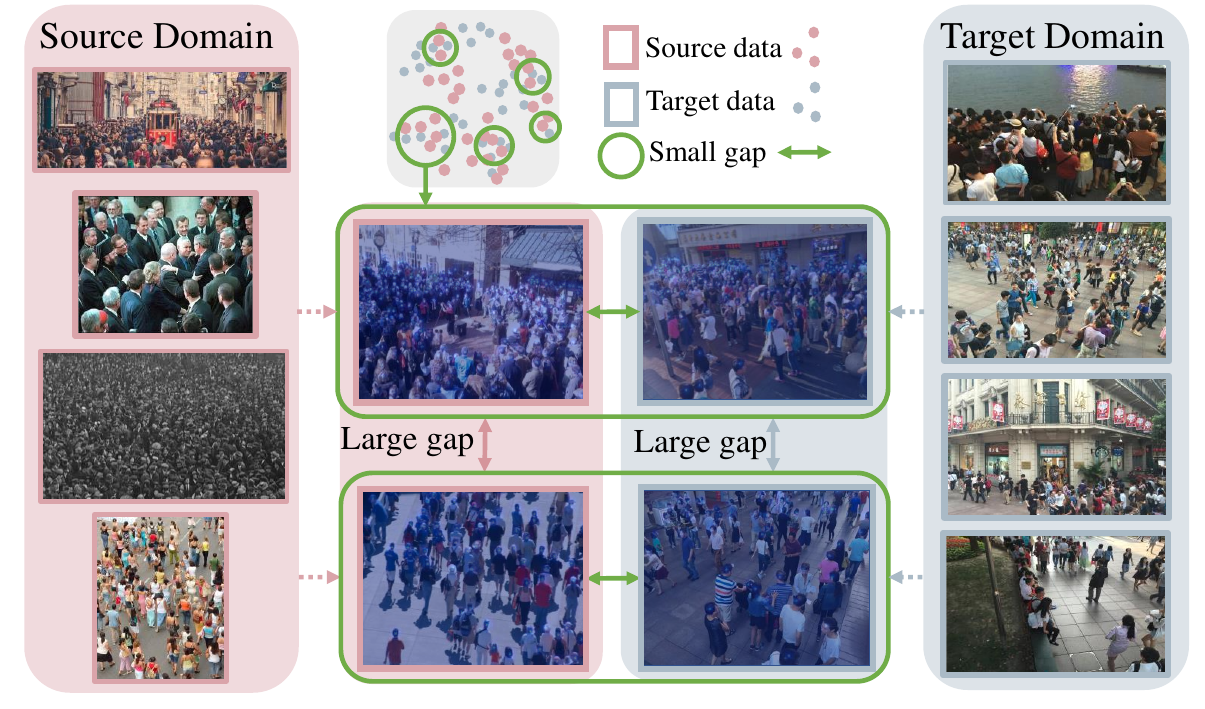}
	\caption{\small Illustration of intra-domain and inter-domain variations in the source (red) and target (blue) domains for factors, \textit{e.g.}, density, scale, and surveillance perspective (domain-agnostic information). The green boxes highlight the small gap between images from different domains, where the inter-domain variations in these factors are smaller than their intra-domain variations.}
	\label{fig:Motivation}
\end{figure}

Crowd counting is a critical task in multimedia and computer vision~\cite{jiang2021degrade, xu2022rank, Zhong2022RainyWA, HanZFZ23, Yang2023WinWB,xie2023striking} that aims to estimate the number of people in surveillance images or videos. Three types of methods are commonly used for crowd counting: detection-based~\cite{arteta2014interactive, wang2021consistency}, regression-based~\cite{bernardo2009bayesian}, and density map-based~\cite{cheng2021decoupled, ma2021spatiotemporal, yan2021crowd, zhou2021locality, 9096602, liang2022focal, lin2022boosting, liu2022reducing}. Detection-based methods are unsuitable for estimating the number of people in dense crowds due to occlusion issues, while regression-based methods are less affected by occlusion but cannot obtain the approximate position distribution of people. The emergence of density map-based methods has addressed the shortcomings of the previous two types of methods and has become the mainstream method for crowd counting. However, most crowd counting methods rely on extensive data annotation and cannot be applied to open scenes. The effect of directly training models with labeled datasets and transferring them to unlabeled scenes is often unsatisfactory due to domain gap issues. As a consequence, unsupervised domain adaptation (UDA) methods~\cite{2019CODA, 2019Learning, gao2021domain, tang2021unsupervised, cho2022part} have been widely applied in crowd counting to bridge the domain gap. 
\begin{figure}
	\centering
	\includegraphics[width = 0.85\columnwidth]{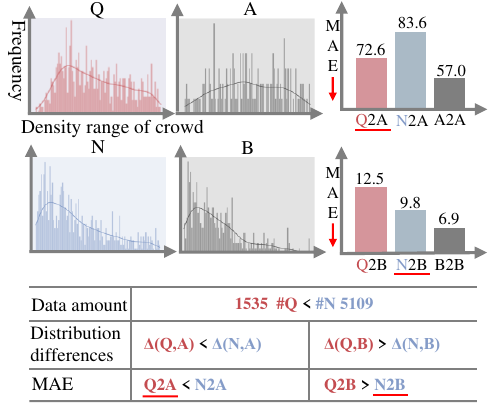}
	\caption{\small Illustration of density distribution and model performance in cross-domain scenarios. Histograms show density proportions of datasets Q, A, N, and B. Performance under different scenarios are displayed in histograms. Tables provide details on data amounts, distribution differences, and cross-domain performance, highlighting the best performance in red. (Baseline model: FIDT~\cite{liang2022focal}.)}
	\label{fig:Motivation2}
\end{figure}
UDA methods in crowd counting cover Synthetic-Real (Syn-Real) or Real-Real scenarios. However, Syn-Real methods often require significant computational resources and cannot capture real crowd characteristics accurately. For the more practical Real-Real cross-domain problem, previous works propose Gaussian process-based iterative learning~\cite{sindagi2020jhu}, and combine detection and density estimation with dual knowledge transfer~\cite{liu2020towards}. Nevertheless, these methods prioritize improving their knowledge and fall short in exploring inter-domain relations. To exploit the inter-domain relation, ASNet~\cite{2021Coarse} use discriminators to evaluate the similarity between the source and target domains at various levels of granularity, while FGFD~\cite{ZhuYYZW22} utilize inter-domain invariant information for model migration. However, as depicted in Fig.~\ref{fig:Motivation}, many factors within the same domain, \textit{e.g.}, density, perspective, and scale, exhibit more significant intra-domain differences than inter-domain differences, limiting the ability to discriminate or find similarities between single images across domains. Some research also focuses on domain generalization, \textit{e.g.}, C2MoT~\cite{wu2021dynamic} and DGCC~\cite{du2022domain}. In C2MoT, only domain-invariant information is explored, but not domain-specific information. In DGCC, although domain information is divided into domain invariant features and domain-specific features, they are considered mutually exclusive and the factual domain-specific features are not explored.

In this paper, we have three key discoveries related to the cross-domain issue in the context of crowd counting. As depicted in Fig.~\ref{fig:Motivation}, factors that are domain-agnostic, \textit{e.g.}, density, perception, and scale, can lead to more significant variations within the same dataset compared to those observed across different domains. This because many of these factors are not unique to a particular domain and can vary widely even within the same dataset. 

Meanwhile, we observe that domain-agnostic factors are misaligned across domains, and the cross-domain performance tends to improve as the different domains become more distributionally consistent. We visualize the distribution of density from {UCF-QNRF (\textbf{Q})}, {NWPU-Crowd} (\textbf{N}), {ShanghaiTech PartB} (\textbf{A}), and {ShanghaiTech PartB} (\textbf{B})~\cite{zhang2016single} at the top of Fig.~\ref{fig:Motivation2}. We discover that the frequency distribution of \textbf{A} and \textbf{Q} are more similar, with a higher frequency of samples with a medium crowd density range. Similarly, the frequency distribution of \textbf{N} and \textbf{B} are also more similar, with a higher frequency of samples with a low crowd density range.
Moreover, we notice that the performance of the model~\cite{liang2022focal} in different cross-domain settings varies significantly. Specifically, the \textbf{N2B} setting outperforms the \textbf{Q2B} setting, and \textbf{Q2A} outperforms \textbf{N2B}, with \textbf{Q2A} even approaching the \textbf{A2A} performance. This discovery indicates that misalignment of domain-agnostic factors at the domain level significantly impacts the model's cross-domain performance~\cite{liang2022focal}.

Furthermore, we discover that the effect of domain misalignment on cross-domain performance can outweigh the impact of the quantity of data available in the source domain.
As shown in the bottom of Fig.~\ref{fig:Motivation2}, the \textbf{Q}~\cite{idrees2018composition} contains 1,535 images, while the \textbf{N}~\cite{wang2020nwpu} contains 5,109 images. However, when the same settings are applied to the target domain \textbf{A}~\cite{zhang2016single}, which consists of 182 images, \textbf{N} performs worse than \textbf{Q}. 
This discovery underscores the criticality of aligning domain-agnostic factors, including compensating for insufficient data quantity, in optimizing cross-domain model performance.

Based on the above three discoveries, we posit that the domain gap is not solely dependent on inter-domain factors at the image level, but is strongly influenced by the misalignment of domain-agnostic factors at the domain level. In this paper, the domain factors can be concretely represented as the crowd distribution, as this encompasses information such as density, perspective, scale, and other related factors.
The objective of this paper is to address domain adaptation by resolving the misalignment of domain-agnostic factors that contribute to the domain gap. Therefore, we propose a three-step domain-adaption approach in an align-and-transfer manner. In the first step, we measure the differences in domain-agnostic factors at the individual level using structural similarity (SSIM) as an indicator. In the second step, we smooth out these domain-agnostic differences. For this purpose, we introduce the optimal transport (OT) strategy to find the optimal domain-to-domain transformation. To handle outlier individuals in the misalignment measurement process, we augment each domain with a virtual column representing a ``dustbin'' individual, where outlier individuals are removed. Finally, we can transfer knowledge based on the aligned domain-agnostic factors. By aligning the domain-agnostic factors across domains through the optimal domain-to-domain transformation, we retrain the model~\cite{liang2022focal} for domain adaptation and bridge the gap across domains.

To summarize, the main contributions are threefold:
\begin{itemize}

	\item We propose a new perspective on the impact of domain-agnostic factors in crowd counting, highlighting their significant influence on in-domain variations and inferior cross-domain performance due to misalignment at the domain level. This discovery provides a better understanding of the domain gap and the importance of domain alignment for improving cross-domain performance.

	\item We design a novel domain adaptation method, called Domain-agnostically Aligned Optimal Transport (DAOT), to bridge the domain gap by aligning domain-agnostic factor misalignments between the source and target domains.

	\item We demonstrate the effectiveness and strong generalizability of the proposed DAOT through extensive experiments on five crowd counting datasets. Experimental results indicate that DAOT outperforms state-of-the-art methods.
\end{itemize}

\section{Related Work}
\subsection{Optimal Transport}
It is well-known that optimal transport (OT)~\cite{peyre2019computational, panaretos2019statistical} has been widely applied in domain adaptation tasks, particularly in generating target domain data by finding the optimal distance between the source and target domain feature distributions. However, most of these methods~\cite{damodaran2018deepjdot, xu2020reliable, 9286483} focus on generating images similar to those in the target domain, which requires finding the optimal distance between the generated and target images. In contrast, the proposed DAOT differs from traditional methods of OT in domain adaptation, as we establish correspondence between source and target domain distributions without generating new images or altering source domain images, similar to a feature-matching task~\cite{jin2021image}.

Compared to point-to-point matching methods~\cite{sarlin2020superglue, han2022dr,wang2023roreg}, the proposed DAOT has a broader scope as it involves region-to-region matching. Moreover, we use a distance measurement method that emphasizes spatial structure, namely SSIM. Note that we do not perform person-level matching, as opposed to DR.VIC~\cite{han2022dr}. Our focus is not on the existence of a matching relation between the query and gallery, but on specific matching and transfer relations.

In crowd counting, OT has been used as a loss function, with DM-Count~\cite{Wang_Liu_Samaras_Nguyen_2020} being one of the first proposed methods. Several methods~\cite{song2021rethinking,lin2023optimal} have adopted this type of loss. Other than loss functions, we use OT to find the best match rather than simply measuring the difference.

\subsection{Unsupervised Domain Adaptation}
Most UDA crowd counting methods~\cite{2019CODA,2019Learning,gao2021domain,liu2023finegrained,xie2023striking} focus on classifying individual images into either the source or target domain, and therefore aim to bridge the gap between the source and target domains on a per-image basis. However, in Real-Real domain adaptation tasks, many images may have greater intra-domain differences than inter-domain ones. Addressing only the differences within individual images may not be sufficient to fundamentally bridge the gap between domains. Instead, we propose a domain adaptation method based on aligning the overall distribution of data in the source and target domains, rather than focusing solely on individual images.
Several crowd counting methods~\cite{2021Coarse, ZhuYYZW22} consider that there are similarities between source and target domains and thus adapt between domains by finding similarities. However, individual similarity retrieval has shifted. Perhaps the found source and target distributions do have strong similarities, but due to the existence of retrieval bias, much of the information in the source distribution is not fully utilized, and only the distribution with the highest similarity is left. In essence, this approach is unable to fully simulate the entire data domain, while the proposed DAOT seeks to maximize the search for transfer relation between the two distributions. 
Apart from the methods that seek to find the similarities between domains, other methods~\cite{wu2021dynamic,du2022domain} attempt to find common knowledge between them. However, in our opinion, many images in different domains differ significantly in density, scale, and perception, making it difficult to find common knowledge. Even within the same domain, it is challenging to find common knowledge. We believe that rich data is essential for the model to learn knowledge and to have stronger generalization abilities. Therefore, instead of seeking common knowledge, we bridge the domain gap by aligning domain-agnostic factor misalignment between the source and target domains.

\begin{figure*}
	\centering
	\includegraphics[width = 0.95\textwidth]{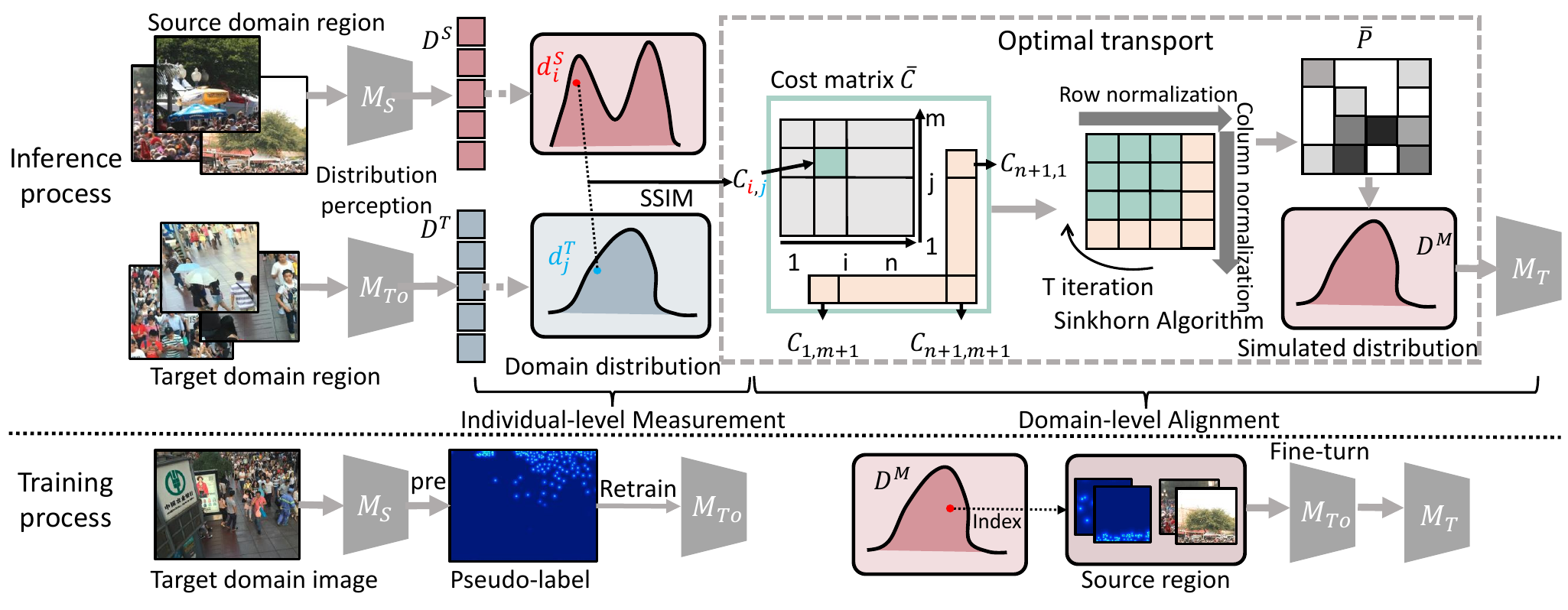} 
	\caption{\small Structure of DAOT. The figure illustrates the inference and training processes, where the training process describes the acquisition of $M_{T0}$ and $M_T$, respectively. The inference process is divided into two stages. In the individual-level measurement stage (stage 1), region information from both the source and target domains is collected. The source domain model $M_S$ and target domain model $M_{T0}$ trained with pseudo-labels from $M_S$ are used for distribution perception, yielding the source domain distribution $D^S$ and target domain distribution $D^T$. The distance matrix $C$ is calculated by SSIM to measure the distance between each distribution and extended to form the cost matrix $\overline{C}$. In the domain-level alignment stage (stage 2), we use the Sinkhorn algorithm with iterative updates to obtain the optimal transfer matrix solution $\overline{P}$ and the final simulated distribution. We fine-tune the initial model $M_{T0}$ using the simulated distribution to obtain the target domain model $M_T$.}
	\label{fig:Framework}
\end{figure*}

\section{Proposed Method}


\subsection{Problem Definition}

In this paper, we consider domain adaptation as a problem of aligning domain-agnostic factors between domains, and we employ a region-level distribution-perception method, where the distribution is used to describe the domain-agnostic factors. As shown in Fig.~\ref{fig:Framework}, the source domain region is set to $S = x_i, i \in (1,n)$, and the target domain region is set to $T = x_j, j \in (1,m)$. The model trained with labeled data in the source domain is defined as $M_S$, the initial model in the target domain is defined as $M_{T0}$, with the training labels using the pseudo-labels predicted by the source domain model on the target domain data set. And the regions in the two domains are distribution-perception to obtain the source domain distribution $D^S = d^S_i$ and the target domain distribution $D^T = d^T_j$, respectively.
In Sec.~\ref{sec:3_1}, the distance matrix $C$ is calculated using SSIM to measure the distance between each distribution and extended to form the cost matrix $\overline{C}$, and the final obtained simulated distribution in Sec.~\ref{sec:3_2} is defined as $D^M = D^S_{r,j}, \{r\} \subseteq \{i\}, i \in (1,n), j \in (1,m)$. The main method for domain alignment is optimal transport (OT).

The principle behind OT is to minimize the cost of transporting one distribution to another, where the cost is usually defined as the distance between the two distributions. The OT problem can be formulated as a linear programming problem, where the objective function is to minimize the cost of transportation subject to some constraints. Therefore, the first step is to obtain a probability transition matrix that expresses the transfer between two distributions, called the cost matrix $\overline{C}$. Then, based on the cost matrix, the Sinkhorn algorithm~\cite{sinkhorn1967concerning, cuturi2013sinkhorn, peyre2019computational} to find the optimal solution $\overline{P}$ solved to find the optimal solution for transporting between the distributions. In our paper, local distribution matching can be achieved by solving an augmented version of Kantorovich's OT problem~\cite{peyre2019computational}. And $U(\overline{A}, \overline{B})$ denotes the discrete of $\overline{A}$ and $\overline{B}$.
\begin{equation}
	L_{\overline{C}}\left(\overline{A},\overline{B}\right) = \min\limits_{\overline{P} \in U(\overline{A}, \overline{B})} \sum_{i,j}^{n,m} \left(\overline{C}_{i,j} \overline{P}_{i,j}\right),
 \label{eq-solve}
\end{equation}
where $\overline{P}$ is a transportation matrix. $\overline{C}_{i,j}$ represents the distance between each element in the source and target domains, in Eq.~\eqref{eq-ssim}.

\subsection{Individual-Level Measurement}
We believe that domain-agnostic factors are variable in both inter-domains and intra-domain, which makes individual images not exclusively belong to a particular domain, while misalignment of these factors at the domain level lead to the domain gap. Therefore, this paper aims to bridge the domain gap by aligning domain-agnostic factors. To achieve domain-level alignment, we measure individual-level differences using the following algorithm.
\label{sec:3_1}

\subsubsection{Describing Domain-Agnostic Factors}
Due to the diversity of distributions increasing with scale, it is challenging to align at the individual level without altering the image itself. Thus, we align the regions in the image to address this issue.
Therefore, we divides the whole image in the source and target domains into different regions and forms the sets $S = x_i$ and $T = y_j$. Subsequently, the distributions $D^S$ and $D^T$ of the source and target domains are obtained after distribution perception of the source domain model $M_S$ and the preliminary target domain model $M_{T0}$. We utilize FIDT~\cite{liang2022focal} as our location-sensitive distribution-perception model.
\subsubsection{Structural Similarity and Cost Matrix}
The solution of the transfer matrix between the distributions heavily relies on the distance metric function between the distributions. In the crowd counting distribution distance metric, due to the huge and dense crowd, it possesses rich brightness, contrast, and structural variation. The SSIM similarity can fully combine this information, and the calculated similarity is not only the value of the vector but the structural properties in the whole distribution. 

For a given source domain distribution set $D^S$ and target domain distribution set $D^T$, we can calculate an SSIM distance, $C_{i,j}$, between each element $d^S_i$ and $d^T_j$. The formula is defined as Eq.~\eqref{eq-ssim}.
\begin{equation}
	C_{i,j} = 1-\text{SSIM} \left(d^S_i,d^T_j\right)
	 = 1-\frac{\left(2 \mu_{d^S_i} \mu_{d^T_j} + Q_1\right) \left(2 \sigma_{d^S_i \cdot d^T_j} + Q_2\right)}{\left(\mu^2_{d^S_i} + \mu^2_{d^T_j} + Q_1\right) \left(\sigma^2_{d^S_i} + \sigma^2_{d^T_j} + Q_2\right)},
	\label{eq-ssim}
\end{equation}
\begin{equation}
	C = \left[
	\begin{array}{lclcl}
	C_{1,1} & \cdots & C_{i,1} & \cdots & C_{n,1}\\
	C_{1,j} & \cdots & C_{i,j} & \cdots & C_{n,j}\\
	C_{1,m} & \cdots & C_{i,m} & \cdots & C_{n,m}\\
	\end{array}
	\right],
	\label{eq-S1}
\end{equation}
where $\mu_{d^S_i}$ and $\sigma_{d^T_j}$ separately denote the mean of $d^S_i$ and standard deviation of ${d^T_j}$. $\sigma_{d^S_i \cdot y_j}$ is covariance of $d^S_i$ and $d^T_j$.
We set $Q_1$ and $Q_2$ as $e{-4}$ and $9e{-4}$. $\overline{C}$ as the final cost matrix, the shape of $(n+1,m+1)$.
\begin{equation}
	\overline{C} = \left[
	\begin{array}{ll}
	C_{i,j} & C_{n+1,1}\\
	C_{1,m+1} & C_{n+1,m+1}\\
	\end{array}
	\right],
	\label{eq-S2}
\end{equation}
where $\overline{C}$ is used to represent the distance between source domain region distribution and target domain region distribution. As shown in Fig.~\ref{fig:ot-c}, to ensure the reliability of the distance scores, we set $C_{i,m+1}$, $C_{n+1,j}$, and $C_{n+1,m+1}$ as ``dustbin'' column to remove unreliable rows and columns, where $C_{i,m+1}$ represents SSIM value between the distribution of each source domain sample and a distribution consisting of all ones, while $C_{n+1,j}$ represents SSIM value between the distribution of each target domain sample and the same distribution consisting of all ones. $C_{n+1,m+1}$ represents the SSIM value between a matrix consisting of all zeros and the same distribution consisting of all ones, which serves as a threshold for determining the reliability of each element in the distance matrix $C$. 
If the minimum value in any row or column of $C$ exceeds $C_{n+1, m+1}$, the corresponding row or column is zeroed out to filter out unreliable elements and enhance the Sinkhorn algorithm's robustness.

\subsection{Domain-Level Alignment}
\label{sec:3_2}

In this paper, the inter-domain adaptation task is transformed into an inter-domain domain-agnostic factor alignment problem, and further, the alignment of domain-agnostic factors is considered as finding the optimal mapping between domain-level distributions, which are used to describe domain-agnostic factors at the domain level. It is also the solution of Eq.~\eqref{eq-solve}. The function we use to find the optimal transport is Sinkhorn~\cite{sinkhorn1967concerning}. 
The ultimate goal of the OT problem is to find a matrix $\overline{P}$ that optimally transfers the source distribution to the target distribution so that the product of $\sum{(\overline{P} \times \overline{C})}$. This is essentially a linear programming problem Eq.~\eqref{eq-solve} with $n + m + 2$ constraints Eq.~\eqref{eq-S2}, the constraints ensure that the probabilities of mapping between the source and target distributions sum up to one for each element, and vice versa.

\subsubsection{Sinkhorn Algorithm}
Sinkhorn Algorithm~\cite{sinkhorn1967concerning,peyre2019computational} is applied to solve for $\overline{P}$, which represents the cost of transporting each element from the source domain to the target domain. The algorithm computes the optimal transport plan from the source domain to the target domain and returns a mapping between the indices of the source and target elements that minimizes the overall cost of the transport plan. In Sec.~\ref{sec:3_2}, the cost matrix $C$, between the domains has been obtained. 

This algorithm is a differentiable version of the Hungarian algorithm~\cite{munkres1957algorithms}, which is typically used for bipartite matching. It involves iterative normalizing $\overline{C}$ along rows and columns, similar to row and column Softmax, and the DAOT for calculating discrete is SSIM. This means minimizing Eq.~\eqref{eq-solve}. Specifically, subject to the constraints that the marginals of $\overline{P}$ match the given weights:

\begin{equation}
	\sum_j \overline{P}_{i,j} = \overline{A}_i, \quad \sum_i \overline{P}_{i,j} = \overline{B}_j,
\end{equation}
where the algorithm updates the rows of $\overline{P}$ as follows:
\begin{equation}
	\overline{A}_i \frac{\overline{B}_j}{\sum_k \overline{P}_{i,k} \overline{C}_{i,k} + \epsilon} \to \overline{P}_{i,j},
\label{eq-iteration}
\end{equation}
where $k$ is an index variable used for summation, it represents that for the $i$-th source sample and the $j$-th target sample, we need to sum over all $k$, which represents the $k$-th sample in the source domain. Therefore, the formula can be understood as computing the transition probability between the $i$-th source sample and the $j$-th target sample and using information from all source domain samples to calculate it. And then updates the columns of $P$ by applying the same operation to the transpose of $P$. The regularization parameter $\epsilon > 0$ prevents division by zero. This row and column scaling process is repeated iteratively until the resulting transport plan $\overline{P}$ converges to a stable solution. The symbol "$\to$" in Eq.~\ref{eq-iteration} indicates the final output of the Sinkhorn Algorithm. The algorithm stops at $T_{\max}=1000$, following the convention in~\cite{cuturi2013sinkhorn}.

\begin{figure}
	\centering
	\includegraphics[width = 1\columnwidth]{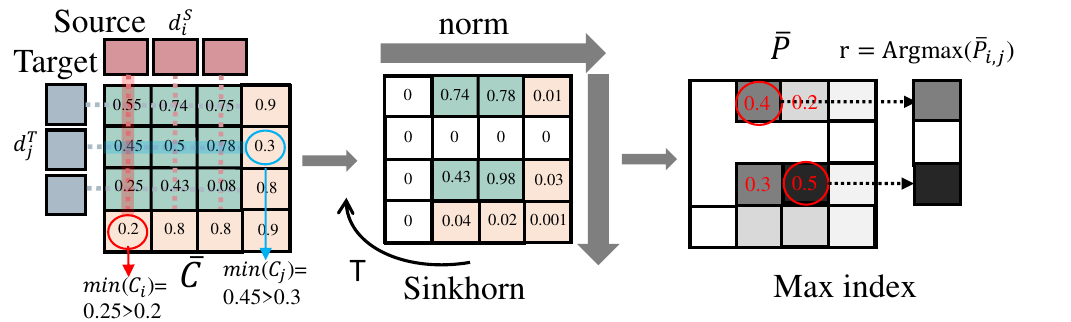}
	\caption{\small Illustration of OT solving process. Red: source domain. Blue: target domain. Green: SSIM distance. Yellow: garbage bin score. Yellow and green form cost matrix $\overline{C}$. $\overline{C}$ regularized iteratively to obtain optimal solution $\overline{P}$, transfer probability matrix. Max value in each row equals final matching distribution index.}
	\label{fig:ot-c}
\end{figure}

\subsubsection{Final Mapping and Target Model}
After obtaining the optimal transport plan $\overline{P}$ using the Sinkhorn algorithm, we can use it to obtain a mapping between the source and target domains. As shown in Fig.~\ref{fig:ot-c}, this mapping represents the correspondence between the source and target samples, and can be used to train a model that can translate samples from the source to the target domain. To obtain the mapping, we find the index $i$ in the source domain that has the highest probability of being paired with each target sample $(d^T_j, d^S_r)$, which can be computed as $r = \arg \max_i \overline{P}_{i,j}$. We consider the correspondences between the source and target samples only if the corresponding probability value is non-zero. This procedure gives us the mapping between the source and target samples with non-zero correspondence values.

Finally, we obtain the distribution $D^M = D^S_{r,j}$ to fine-tune the initial model $M_{T0}$ and obtain the final target domain model $M_T$.

\begin{table*}
	\centering
	\caption{\small Evaluation performance comparison of MAE and RMSE with the state-of-the-arts for cross-domain crowd counting. ``Syn'' and ``Real'' indicate the synthetic GCC dataset and the real datasets as source domains, respectively. ``AD'' stands for Domain Adaption. $\dag$ indicates reproduced results. Bold and underline numbers denote the best and second-best results, respectively.}%
	\footnotesize
	\setlength{\tabcolsep}{3.1pt}
	\begin{tabular}{lccccccccccccccccccc}
	\toprule
	\multirow{2}[2]{*}{Method} & \multirow{2}[2]{*}{backbone} &
	\multirow{2}[2]{*}{Type} & 
	\multirow{2}[2]{*}{AD}
	 & \multicolumn{2}{c}{A2B} 
	 & \multicolumn{2}{c}{A2Q}
	 & \multicolumn{2}{c}{Q2A}
	 & \multicolumn{2}{c}{Q2B}
	 & \multicolumn{2}{c}{N2A}
	 & \multicolumn{2}{c}{N2B}
	 & \multicolumn{2}{c}{J2B}
	 & \multicolumn{2}{c}{J2Q}\\
	\cmidrule(lr){5-6} \cmidrule(lr){7-8} \cmidrule(lr){9-10} \cmidrule(lr){11-12} 
	\cmidrule(lr){13-14} \cmidrule(lr){15-16} \cmidrule(lr){17-18} \cmidrule(lr){19-20}
	 & & & & MAE & RMSE & MAE & RMSE & MAE & RMSE & MAE & RMSE & MAE & RMSE & MAE & RMSE & MAE & RMSE & MAE & RMSE\\
	\midrule
 	Cycle GAN~\cite{zhu2017unpaired} & CycleGAN & Syn & \CIRCLE & 25.4 & 39.7 & 257.3 & 400.6 & 143.3 & 204.3 & 25.4 & 39.7 & 143.3 & 204.3 & 25.4 & 39.7 & 25.4 & 39.7 & 257.3 & 400.6\\
	GP~\cite{sindagi2020jhu} & ResNet50 & Syn & \CIRCLE & 12.8 & 19.2 & 210.0 & 351.0 & 121.0 & 181.0 & 12.8 & 19.2 & 121.0 & 181.0 & 12.8 & 19.2 & 12.8 & 19.2 & 210.0 & 351.0\\
	DACC~\cite{gao2021domain} & CycleGAN & Syn & \CIRCLE & 13.1 & 19.4 & 203.5 & 343.0 & 112.4 & 176.9 & 13.1 & 19.4 & 112.4 & 176.9 & 13.1 & 19.4 & 13.1 & 19.4 & 203.5 & 343.0\\
 	BLA~\cite{gong2022bi} & VGG16 & Syn & \CIRCLE & 11.9 & 18.9 & 198.9 & 316.1 & ~~99.3 & {145.0} & 11.9 & 18.9 & ~~{99.3} & {145.0} & 11.9 & 18.9 & 11.9 & 18.9 & 198.9 & 316.1\\
 	CDCC~\cite{liu2022leveraging} & CycleGAN & Syn & \CIRCLE & 11.4 & \textbf{17.3} & 198.3 & 332.9 & 109.2 & 168.1 & 11.4 & \textbf{17.3} & 109.2 & 168.1 & 11.4 & {17.3} & 11.4 & \textbf{17.3} & 198.3 & 332.9\\
	\midrule
	SPN+L2SM~\cite{xu2019learn} & VGG16 & Real & \Circle & 21.2 & 38.7 & 227.2 & 405.2 & ~~73.4 & {119.4} & - & - & - & - & - & - & - & - & - & -\\
	BL~\cite{ma2019bayesian} & VGG16 & Real & \Circle & 15.9 & 25.8 & 135.8 & 252.6 & ~~69.8 & 123.8 & 15.3 & 26.5 & - & - & - & - & - & - & - & -\\
	RBT~\cite{liu2020towards} & VGG16 & Real & \CIRCLE & 13.4 & 29.3 & 175.0 & 294.8 & - & - & - & - & - & - & - & - & - & - & - & -\\
	D2CNet~\cite{cheng2021decoupled} & VGG16 & Real & \CIRCLE & 21.6 & 34.6 & 126.8 & 245.5 & ~~\textbf{61.2} & \textbf{100.8} & \textbf{10.2} & \textbf{19.1} & ~~93.3 & 167.1 & 11.1 & \underline{16.0} & 27.4 & 43.2 & 171.4 & 302.2\\
 	ASNet~\cite{2021Coarse} & VGG16 & Real & \CIRCLE & 13.6 & 23.2 & - & - & - & - & - & - & - & - & - & - & - & - & - & -\\
 	C2MoT~\cite{wu2021dynamic} & VGG19 & Real & \CIRCLE & \underline{12.4} & \underline{21.1} & 125.7 & 218.3 & - & - & - & - & - & - & - & - & - & - & \textbf{135.3} & \textbf{251.9}\\
 	FGFD~\cite{ZhuYYZW22} $\dag$ & HRNet & Real & \CIRCLE & 12.7 & 23.3 & 124.1 & 242.0 & ~~70.2 & 118.4 & 12.5 & {20.3} & ~~\underline{83.2} & \underline{138.1} & 10.2 & 18.6 & \underline{13.4} & \underline{20.9} & 166.7 & 320.0\\
	DGCC~\cite{du2022domain} & VGG16 & Real & \CIRCLE & 12.6 & 24.6 & \underline{119.4} & \underline{216.6} & ~~67.4 & \underline{112.8} & 12.1 & 20.9 & - & - & - & - & - & - & - & -\\
	\midrule
	FIDT~\cite{liang2022focal} $\dag$ & HRNet & Real & \Circle & 16.0 & 26.9 & 129.3 & 254.1 & ~~72.6 & 132.3 & 12.5 & 21.7 & ~~83.6 & \underline{138.1} & ~~\underline{9.8} & 17.0 & 14.2 & 27.3 & 166.6 & 282.4\\
	DAOT (Ours) & HRNet & Real & \CIRCLE & \textbf{10.9} & \textbf{18.8} & \textbf{113.9} & \textbf{215.6} & ~~\underline{67.0} & 128.4 & \underline{11.3} & \underline{19.6} & ~~\textbf{79.9} & \textbf{135.0} & ~~\textbf{8.8} & \textbf{14.8} & \textbf{10.5} & \textbf{17.4} & \underline{146.5} & \underline{255.2}\\
	\bottomrule
	\end{tabular}
	\label{tab:ExpSOTA}
\end{table*}%

\section{Experimental Result}
\subsection{Datasets and Implementation Details}

\textbf{Shanghai Tech}~\cite{zhang2016single} is a benchmark for crowd counting contains 1,198 scene images and 330,165 labeled head positions, which is subdivided into PartA (\textbf{A}) and PartB (\textbf{B}).
\textbf{A} is densely populated and contains 300 training images and 182 testing images; \textbf{B} is relatively sparser and contains 400 training images and 316 testing images. 

\textbf{UCF-QNRF} (\textbf{Q})~\cite{idrees2018composition} is a challenging crowd counting dataset with rich scenes and varying viewpoints, densities, and illumination conditions. It contains 1,535 images of dense crowd scenes, with 1,201 training images and 334 testing images.

\textbf{NWPU-Crowd} (\textbf{N})~\cite{wang2020nwpu} has 5,109 high resolution images and annotated entities 2,133,238. The large range of annotated entity numbers for a single image is extremely challenging. Additionally, some images have object annotation number of 0.

\textbf{JHU-CROWD++} (\textbf{J})~\cite{sindagi2020jhu} contains 4,372 images with a total of 1.51 million annotations, including some based on extreme weather variation and light variation, and introduces negative samples of entities annotated with zero.

\textbf{Loss Function.}
We use the loss function proposed in FIDT~\cite{liang2022focal} for training, where $\eta$ is set to 1 following FIDT. 
\begin{equation}
	\mathcal{L} = \mathcal{L}_{MSE} + \eta \mathcal{L}_{I-S}.
	\label{eq-L}
\end{equation}

\textbf{Evaluation Metrics.}
In this paper, the evaluation metrics used are mean absolute error (MAE) and root mean square error (RMSE) introduced in~\cite{zhang2016single}. MAE assesses model accuracy, while RMSE evaluates model robustness.

\textbf{Optimal Transport.}
In \textbf{A2Q} setting, the regularization parameter $\epsilon$ is set to $e{-2}$, while in \textbf{Q2B} setting, it is set to $e{-1}$. All other settings are set to $e{-3}$. The algorithm stops at $T_{\max}=1000$.

\textbf{Retraining.}
We use adaptive moment estimation (Adam)~\cite{KingmaB14} to optimize the model with the learning rate of $e{-5}$, and the weight decay is set as $5e{-4}$. We set the size of the training batch to $16$. 

\subsection{Comparison with the State-of-the-Arts}

To evaluate the effectiveness of DAOT, this subsection compares it with other state-of-the-art cross-domain methods. We conduct quantitative evaluations through five sets of adaptation experiments, reporting the error between the predicted density maps and ground-truth maps on the target set for each dataset pair. We compare DAOT with three types of counting methods:
1) state-of-the-art counting methods without domain adaptation~\cite{xu2019learn, cheng2021decoupled},
2) Syn-Real domain adaptation methods~\cite{zhu2017unpaired,2019Learning,sindagi2020jhu,gao2021domain,gong2022bi,liu2022leveraging}, and
3) Real-Real domain adaptation methods~\cite{2021Coarse,wu2021dynamic,du2022domain,ZhuYYZW22}, as well as our baseline method, FIDT~\cite{liang2022focal}. Notably, the source domains for Syn-Real methods are large-scale synthetic GCC~\cite{2019Learning}. 

As shown in Table~\ref{tab:ExpSOTA}, the proposed DAOT achieves the best performance in five groups' experiments (\textbf{A2B}, \textbf{A2Q}, \textbf{N2A}, \textbf{N2B}, and \textbf{J2B}) and obtained the second-best performance among the other groups (\textbf{Q2A}, \textbf{Q2B}, and \textbf{J2Q}). Especially with a significant improvement over other methods in the setting of \textbf{A2B} and \textbf{N2B}. The DAOT outperforms the baseline in all 8 settings and achieves better results than the similarity-based FGFD~\cite{ZhuYYZW22} method in multiple settings. 
Compared with D2CNet~\cite{cheng2021decoupled}, the proposed DAOT is slightly lower than D2CNet, indicating that D2CNet is more suitable for the group with the source domain as \textbf{Q}, while the DAOT has a stronger generalization ability. Compared with C2MoT~\cite{wu2021dynamic}, the DAOT is slightly lower than C2MoT, but the amount of data used for retraining is much lower than that of C2MoT, and C2MoT is more dependent on large amounts of data. When the target domain is \textbf{B}, DAOT often achieves better performance, which proves that it has a stronger fitting ability for sparse density distribution. In the setting of \textbf{A2Q} and \textbf{N2B} with aligned domain-agnostic factors, the proposed DAOT method can further improve the excellent performance.
\begin{figure}
	\centering
	\includegraphics[width = 0.99\columnwidth]{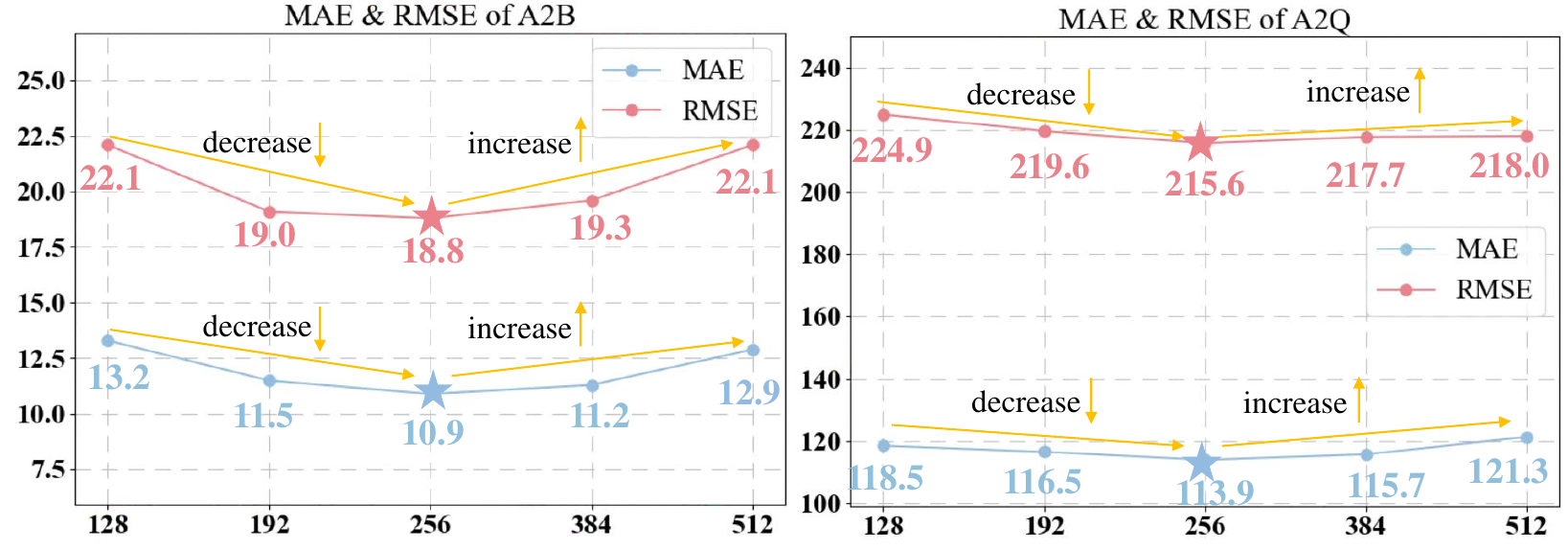}
	\caption{\small Line plot depicting the effect of region size on cross-domain performance, with stars indicating optimal values.}
	\label{fig:region}
\end{figure}
\begin{figure*}
	\centering
	\includegraphics[width = 0.95\textwidth]{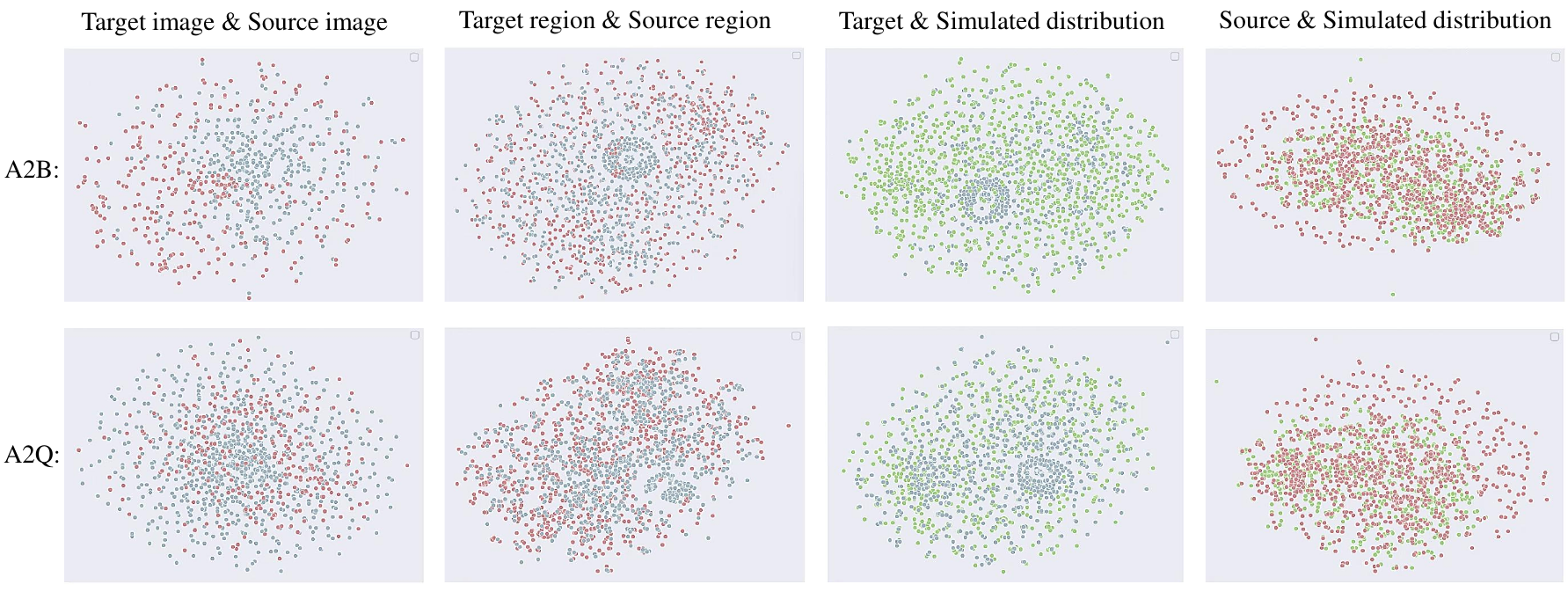} 
	\caption{\small Visualization of t-SNE for different distributions. The two rows respectively showcase the clustering results of A2B and A2Q, where blue represents the target domain distribution, red represents the source domain distribution, and green represents the simulated distribution. The four columns illustrate the image distribution of the target and source domains, the distribution of the regions in the target and source domains, the clustering of the target domain and simulated distribution, and the clustering of the source domain and simulated distribution.}
	\label{fig:t-SNE}
\end{figure*}

\subsection{Ablation Study}

\textbf{Ablation Study on Region Size}
The transition relation between entire images across domains is limited, while the transition relation between regions exists extensively across different domains. As the region size decreases, the accuracy of distance measurement between distributions will decrease. As the region size increases, the transition relation between distributions gradually decreases. As shown in Fig.~\ref{fig:region}, as the region size decreases from the entire image to 256, the MAE and RMSE metrics correspondingly improve. However, as the region size continues to decrease beyond 256, MAE and RMSE begin to exhibit a rising trend.

\textbf{Ablation Study on Various Factors Performance}
To achieve domain adaptation in crowd counting, we transform it into a domain-agnostic factors alignment problem and adopt optimal transport (OT) to find the best transfer distance between distributions.
In Table~\ref{tab:abalation_factors}, using all factors achieves the best performance in all experiments. OT significantly improved performance, especially in \textbf{Q2A} experiment. Using only pseudo-labels and SSIM achieves the second-best performance, showing a slight improvement over using only source-domain data. This suggests that adjusting the density distribution misalignment can bridge the domain gap.

\begin{table}
	\centering
 	\caption{\small Ablation studies of diverse factors on cross-domain performance. Each row exhibits the MAE of models trained with only source domain data, pseudo-labels generated by predicted density maps (Pre.), SSIM, and OT, respectively.} 
	\footnotesize
	\setlength{\tabcolsep}{1.9pt}
	\begin{tabular}{cccccccccccc}
	\toprule
	\multirow{2}[2]{*}{Source} & \multirow{2}[2]{*}{Pre.} & \multirow{2}[2]{*}{SSIM} & \multirow{2}[2]{*}{OT} & \multicolumn{2}{c}{A2B} & \multicolumn{2}{c}{A2Q} & \multicolumn{2}{c}{Q2A} & \multicolumn{2}{c}{Q2B}\\
 	\cmidrule(lr){5-6} \cmidrule(lr){7-8}\cmidrule(lr){9-10}\cmidrule(lr){11-12}
	 & & & & MAE & RMSE & MAE & RMSE & MAE & RMSE & MAE & RMSE\\
	\midrule
	\CIRCLE & \Circle & \Circle & \Circle & 16.0 & 26.9 & 129.3 & 254.1 & 72.6 & 132.3 & 12.5 & 21.7\\
 	\Circle & \CIRCLE & \Circle & \Circle & 13.1 & 27.1 & 120.5 & 216.8 & 76.9 & 138.4 & 13.7 & 22.6\\
	\Circle & \CIRCLE & \CIRCLE & \Circle & 11.0 & 19.6 & 116.5 & 219.6 & 76.4 & 138.1 & 12.3 & 24.3\\ 
	\Circle & \CIRCLE & \CIRCLE & \CIRCLE & \textbf{10.9} & \textbf{18.8} & \textbf{113.9} & \textbf{215.6} & \textbf{67.0} & \textbf{128.4} & \textbf{11.3} & \textbf{19.6}\\
	\bottomrule
	\end{tabular}
	\label{tab:abalation_factors}
\end{table} %
\begin{table}
	\centering
	\caption{\small Ablation studies of distance metrics on cross-domain performance. Each row shows the performance using Euclidean distance (ED), Kullback-Leibler (KL) divergence, and SSIM.} 
	\footnotesize
 	\setlength{\tabcolsep}{4.4pt}
	\begin{tabular}{lcccccccc}
	\toprule
	\multirow{2}[2]{*}{Metric} & \multicolumn{2}{c}{A2B} & \multicolumn{2}{c}{A2Q} & \multicolumn{2}{c}{Q2A} & \multicolumn{2}{c}{Q2B}\\
	\cmidrule(lr){2-3} \cmidrule(lr){4-5}\cmidrule(lr){6-7}\cmidrule(lr){8-9}
	 & MAE & RMSE & MAE & RMSE & MAE & RMSE & MAE & RMSE\\
	\midrule
	Source &16.0 & 26.9 & 129.3 & 254.1 & 72.6 & 132.3 & 12.5 & 21.7\\
	ED & 14.4 & 25.3 & 125.6 & 218.3 & 68.1 & 130.0 & 14.1 & 25.9\\
	KL & 11.2 & 21.0 & 114.7 & 218.9 & 80.7 & 150.3 & 13.3 & 27.0\\
	SSIM & \textbf{10.9} & \textbf{18.8} & \textbf{113.9} & \textbf{215.6} & \textbf{67.0} & \textbf{128.4} & \textbf{11.3} & \textbf{19.6}\\ 
	\bottomrule
	\end{tabular} 
	\label{tab:abalation_metric}
\end{table} %

\textbf{Ablation Study on Distance Metric}
This paper aims to find the optimal transport between distributions, by measuring the distribution distance between the source and target domains before using OT. The OT ultimately obtains a more reasonable distribution distance. The distance measurement method is crucial for the final solution, and we conduct ablation studies on different similarity metrics, which are shown in Table~\ref{tab:abalation_metric}. The three distance metric methods have all shown some improvement compared to using source domain data alone, especially when using SSIM.

\subsection{Qualitative Analysis}

\textbf{Analysis of t-SNE Visualization}
To more intuitively demonstrate the changes in domain-agnostic factors before and after domain adaptation, we visualize Fig.~\ref{fig:t-SNE}. In \textbf{A2B}, the clustering of the overall image distribution shows the most obvious difference, with the distribution of \textbf{B} more concentrated in the lower left corner. This indicates that there is more similarity in the crowd distribution between \textbf{A} and \textbf{B}, but the range of \textbf{A} is wider. In the regional clustering, most of the data in the regions of \textbf{A} and \textbf{B} overlap, but the distribution range of \textbf{A} is still wider than that of \textbf{B}. From the third and fourth columns, it can be seen that although the simulated distribution comes entirely from the source domain, it is highly overlapped with the target domain and more concentrated than the distribution of the source domain.

In \textbf{A2Q}, the clustering of the overall image distribution shows higher overlap, but the distribution of \textbf{Q} is more concentrated in the center. In regional clustering, the distribution of \textbf{Q} shows a more obvious clustering phenomenon. In the visualizations of the third and fourth columns, it can be clearly seen that the simulated distribution is closer to the target domain.

\begin{figure*}
	\centering
	\includegraphics[width = 0.91\textwidth]{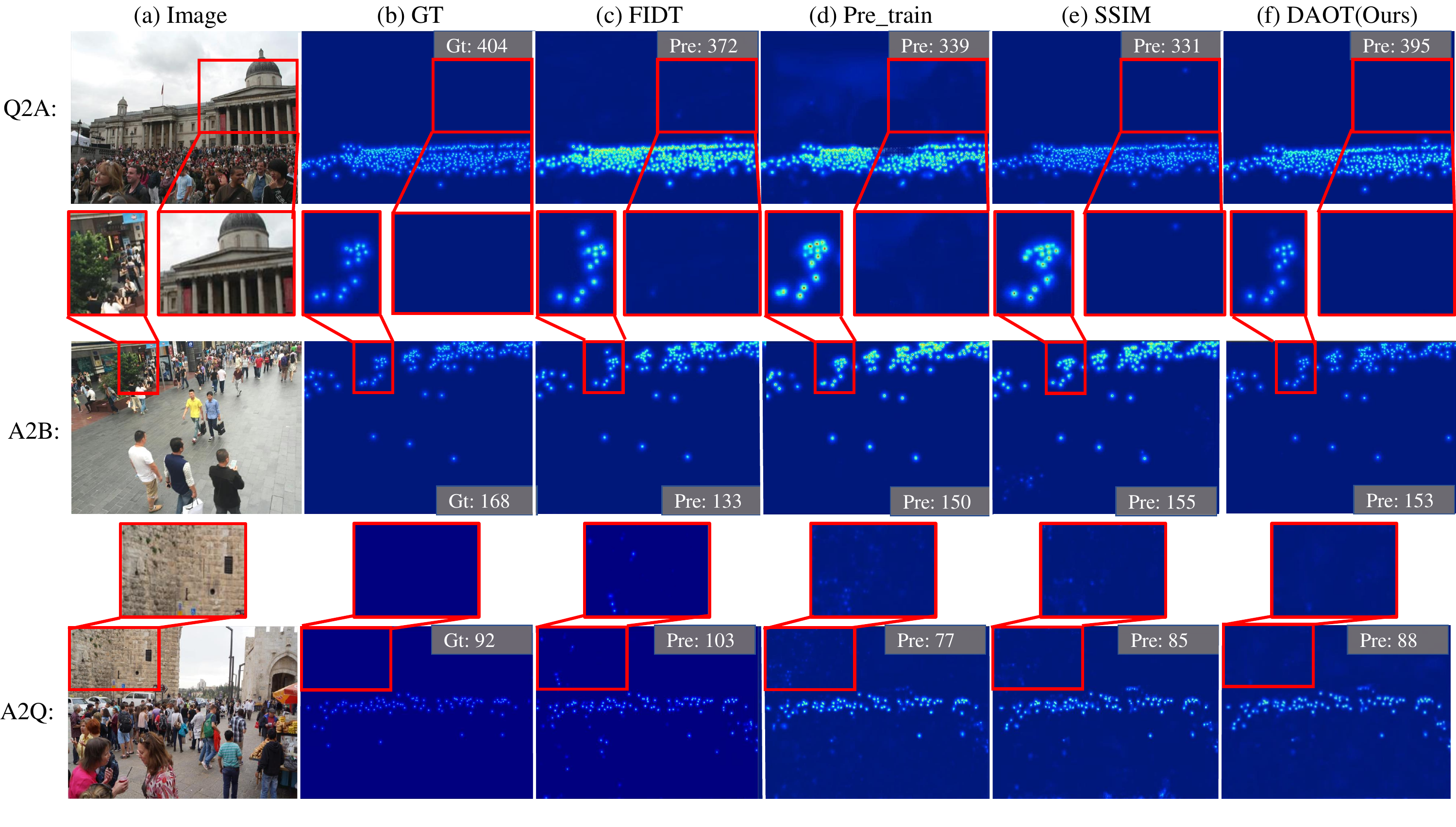}
	\caption{\small Visualization results in cross-domain setting involving Q2A, A2B, and A2Q. We demonstrate cross-scene results of baseline FIDT~\cite{liang2022focal} (c), the method of using the target domain predicted density map as a pseudo label (d), the method only based on SSIM (e), as well as DAOT (Ours) (f). Our DAOT yields superior visual results and more accurate counting consequences than others.}
	\label{fig8}
\end{figure*}

\textbf{Analysis of Cross-Domain Density Map Prediction}
To visualize the cross-domain performance of different methods in a straightforward manner, we visualize Fig.~\ref{fig8}.
The figure illustrates that the proposed DAOT method outperforms other methods in various scenarios, ranging from dense to sparse (\textbf{B2A}), from large-scale to small-scale (\textbf{Q2A}), and even from small-scale dense to large-scale (\textbf{A2Q}). Particularly in \textbf{Q2A} setting, the DAOT prediction is very close to the ground truth. In the first two rows (\textbf{Q2A} and \textbf{A2Q}), there are fewer erroneous classifications of the background, and in the third row (\textbf{A2B}), the estimation of the crowd is more accurate. We have highlighted these parts with red boxes.

\begin{figure}
	\centering
	\includegraphics[width = 0.9\columnwidth]{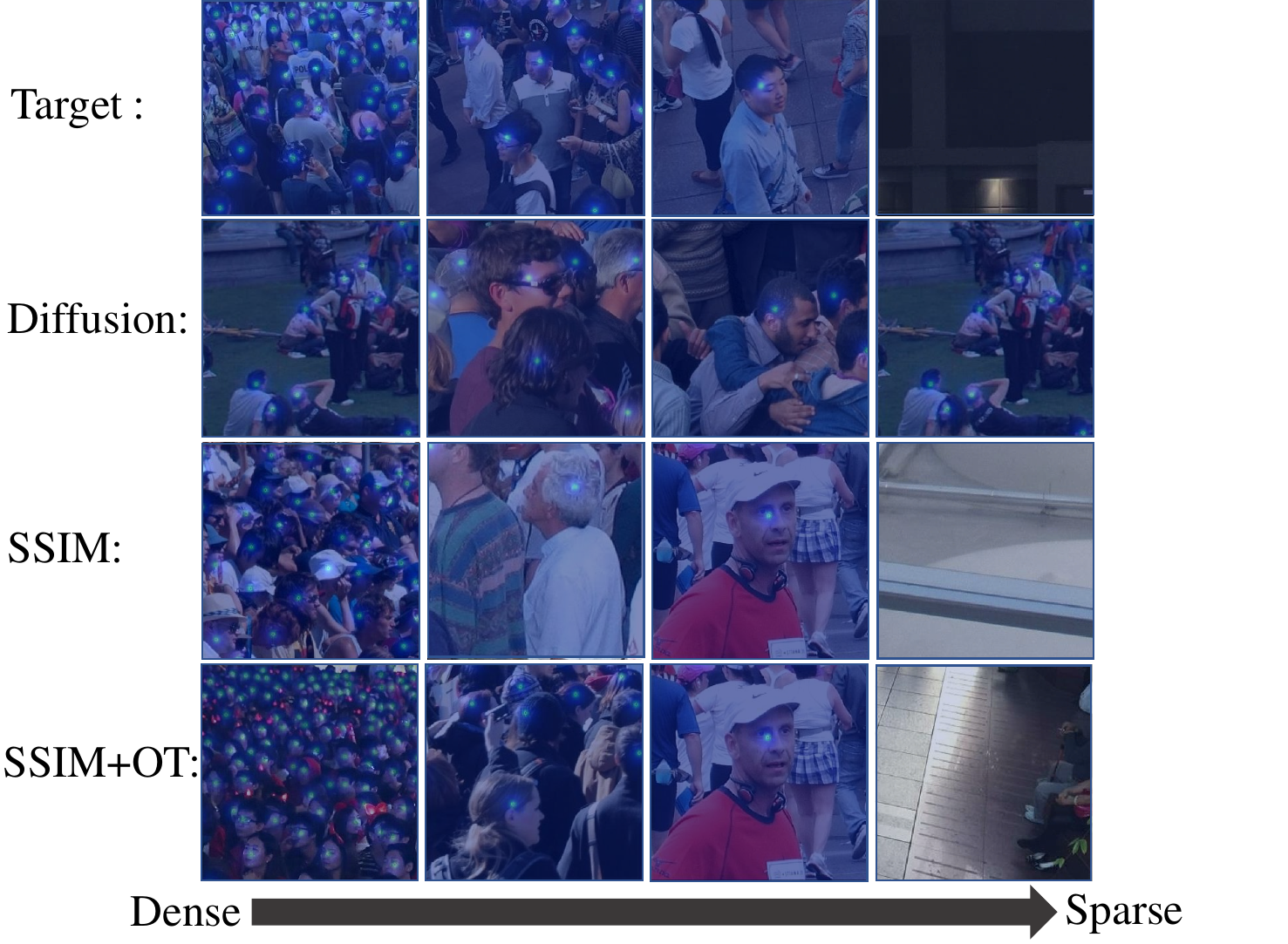}
	\caption{\small Visualization of patches retrieved from source domain to target domain of varying densities in Q2B. Each row corresponds to the best-matching region retrieved by Diffusion, SSIM-based, and SSIM + OT (Ours), respectively, with a region size of 256 pixels.}
	\label{fig:match}
\end{figure}

\textbf{Analysis of Distribution Match}
The final output distribution has a matching relation with the initial target domain distribution, and we visualize the distribution matching in Fig.~\ref{fig:match}. We demonstrate the retrieved source domain regions using different methods on the target domain regions ranging from dense to sparse. As shown in the figure, the DAOT achieves superior performance across all density levels, even in scenarios with few people or only background. Compared with the method using only SSIM, the proposed DAOT achieves better performance in the second column with denser scenarios, and has better perception simulation for the densest scenario. Compared with the diffusion-based FGFD~\cite{ZhuYYZW22}, it achieves better results in the densest and sparsest scenarios, and has better simulation for the location of pedestrians.

\section{Conclusion}
In crowd counting domain adaptation task, it has been observed that the intra-domain differences are larger than the inter-domain differences in many data domains. However, even for large-scale source domains, the model's performance in cross-domain tasks is consistently inferior to that in the source domain. Our research has revealed that this is mainly due to the misalignment of domain-agnostic factors across domains, which strongly influences the cross-domain performance of the model. Therefore, we propose to bridge the inter-domain gap in crowd counting by aligning the misalignment of domain-agnostic factors using optimal transportation in both the source and target domains. Our approach is the first domain adaptation method for crowd counting that considers the entire data domain, unlike previous methods that deal with individual distributions. We treat the overall domain distribution as a whole and achieve the best performance on multiple datasets. This work provides a new perspective on crowd counting domain adaptation, and we believe that solving the problem of domain misalignment will benefit not only cross-domain tasks but also other related applications. 

\begin{acks}
This work was supported in part by the National Natural Science Foundation of China (62271361 and 62171325), and the Department of Science and Technology, Hubei Provincial People’s Government (2021CFB513), Guangdong Distinguished Young Scholar (2023B1515020097), and Singapore MOE Tier 1 Fund (MSS23C002).
\end{acks}
\newpage
\bibliographystyle{ACM-Reference-Format}
\bibliography{Counting}

\appendix
\section*{Appendix}
\section{Experiment on Virtual Data}
To demonstrate the significant impact of the misalignment between the source and target domains on the cross-domain performance, we simulated a dataset (from A and Q) similar to the A test set distribution as the training set and evaluated it on the test set. 

\begin{figure}[!h]
	\centering
	\includegraphics[width = 0.85\columnwidth]{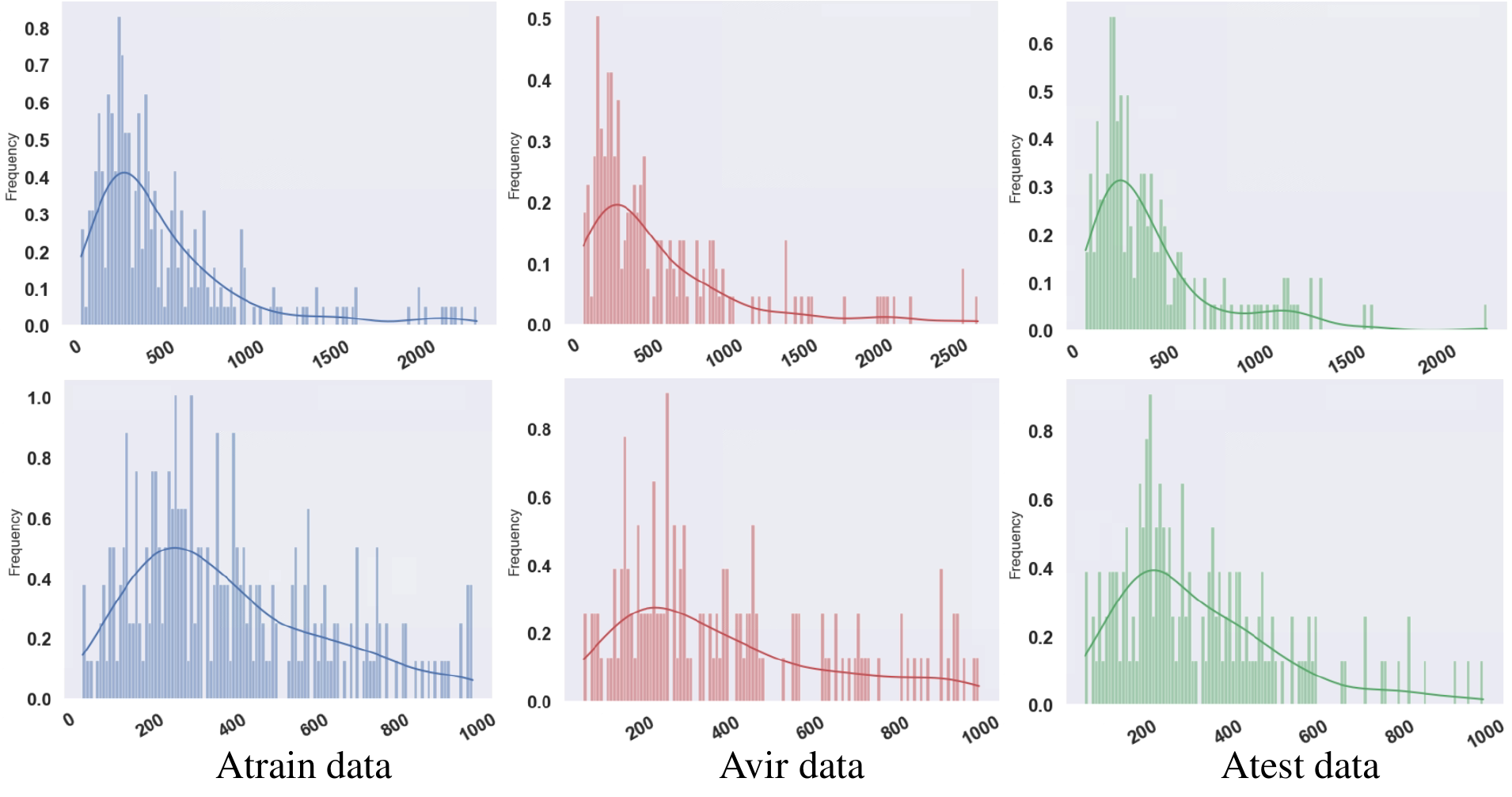}
	\caption{\small Histograms of crowd density frequency for the training set (blue), test set (green), and virtual dataset (red, simulated data with the same distribution as the test set) of Dataset A, with a group setting of 100. The first row shows all data, while the second row shows data from 0 to 1000.}
	\label{fig:fre}
\end{figure}

As shown in Fig.~\ref{fig:fre} and Table~\ref{tab:abalation_}, when the crowd distribution in the training data is similar to that in the test set, the model's performance is also close to the non-cross-domain performance. Even in cases where the data distribution is closer, less data has achieved better performance.

\begin{table}[!h]
	\centering
	\caption{\small Comparison of performance between virtual data training and target domain training set training.}
	\small 
	\begin{tabular}{lcccccc}
	\toprule
	{Setting} & {A2A} & {V2A} \\
	\midrule
	 Amount of Training Data & 300 & 295\\
	MAE & 70.9 & 67.0\\
	RMSE & 140.7 & 134.0\\
	\bottomrule
	\end{tabular}
	\label{tab:abalation_}
\end{table}%
\begin{figure}
	\centering
	\includegraphics[width = 0.98\columnwidth]{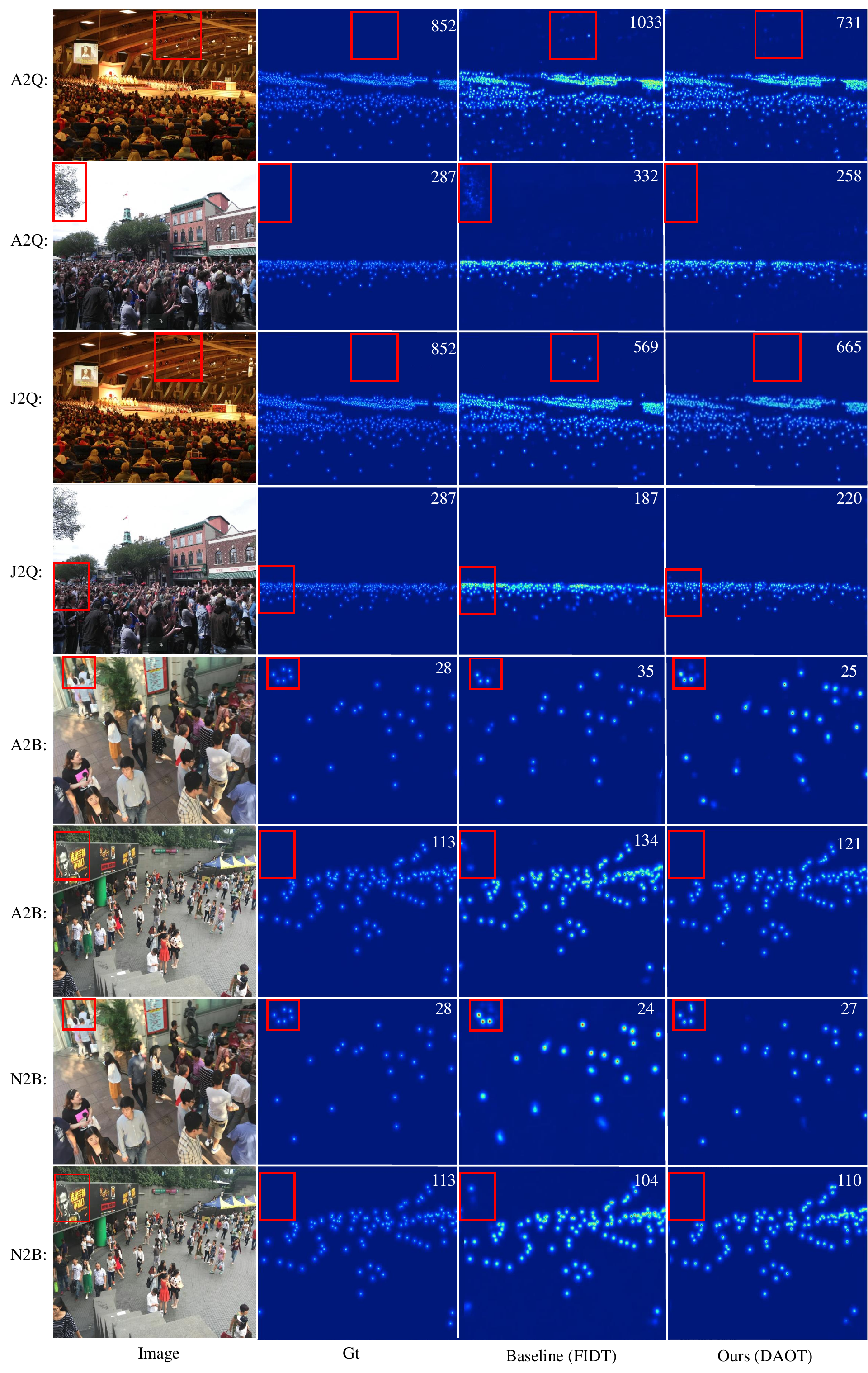}
	\caption{\small Visualization results of density map. The density prediction maps for A2Q, J2Q, A2B, and N2B are displayed. Each column represents the image, ground truth (GT), baseline (FIDT), and our method (DAOT), respectively.}
	\label{fig:density map}
\end{figure}

\begin{figure*}
	\centering
	\includegraphics[width = 0.75\textwidth]{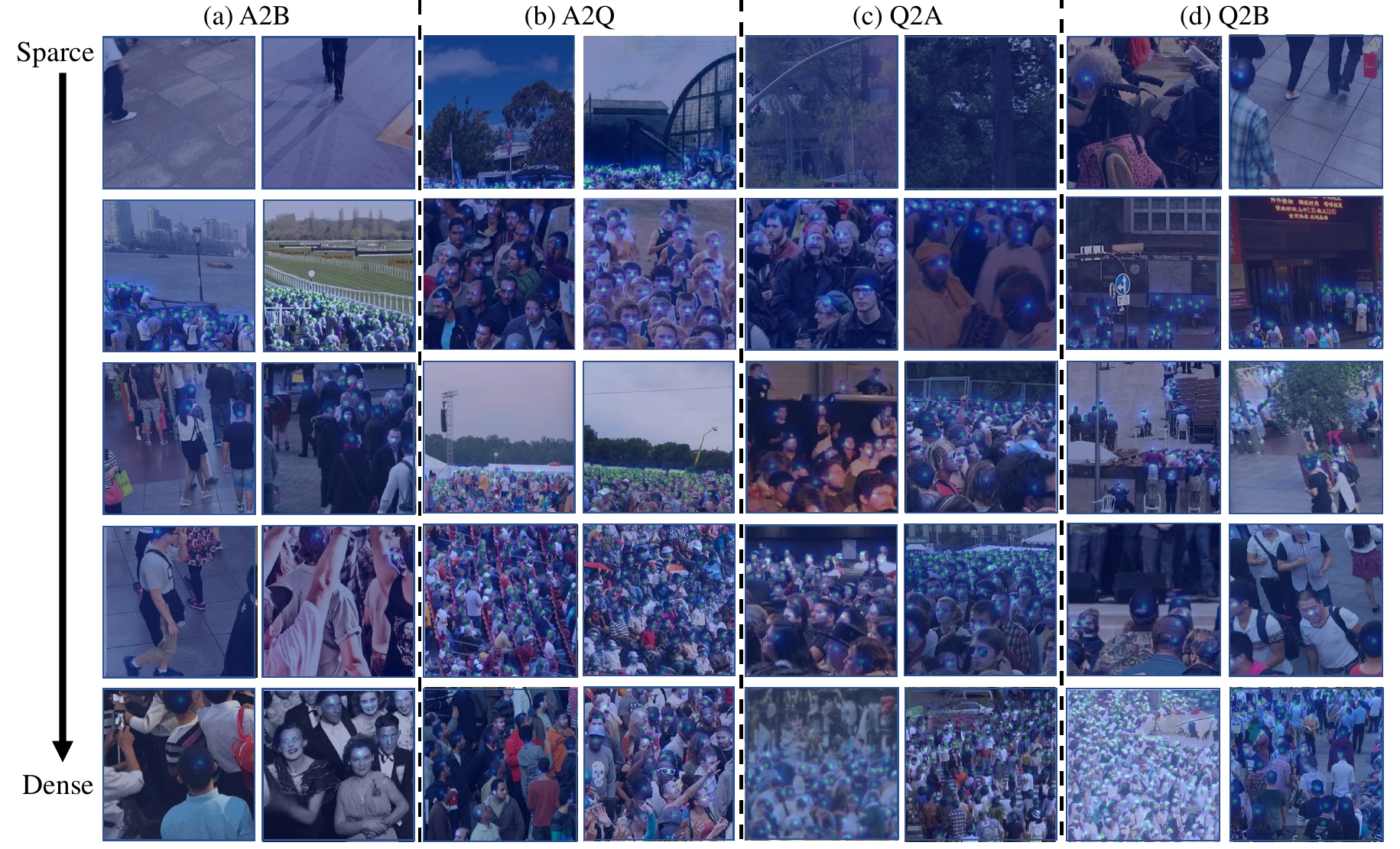}
	\caption{\small Visualization results in distribution match. In the four sets of cross-domain experiments A2B, A2Q, Q2A, and Q2B. The shown patches come from the source and target domains and are arranged from top to bottom according to different crowd density levels.}
	\label{fig:match}
\end{figure*}
\section{Visualization of density map}
We show in Fig.~\ref{fig:density map} the detailed visualization of the cross-domain experiments on A2Q, J2Q, A2B, and N2B. Each column stands for the crowd image, the corresponding Ground truth, the density maps produced by the Baseline method FIDT and the results predicted by our DAOT method.
Our method has been demonstrated to outperform the baseline approach in scenes with varying crowd densities, including dense-to-sparse (A2B), sparse-to-dense (A2Q), and dense-to-dense (J2Q, N2B) scenes. The primary reason is that our method leverages the alignment of domain-agnostic factors between the source and target domains, facilitating effective model migration. Specifically, our method is able to learn to match the density, surveillance perspective, scale, and other relevant factors in a domain-agnostic way, which improves the ability of the model to generalize to new domains with different crowd distributions. This lets the model perceive the distribution and background better, leading to more accurate crowd counting.
\section{Visualization of various factors alignment}

We visualize the distribution matching results in Fig.~\ref{fig:match} under different settings, arranged from sparse to dense from top to bottom. Our method aligns the distributions in terms of density, head scale, and perception in each group, as demonstrated by the visualizations in Fig.~\ref{fig:match}, which are arranged from sparse to dense. Notably, in the third row of A2Q, the crowds in the two patches on the left and right are extremely similar in density, scale, perception, and even background factors. Additionally, the first row of A2B and Q2A shows similar background matching in the absence of people. These results demonstrate that our method can align various factors of crowd distributions.

\section{Regularization Term}
The OT regularization term enhances the smoothness and stability of the OT plan by penalizing the cost function. It usually measures the distance or divergence between the transport plan and a reference distribution, often a prior distribution. This regularization prevents overfitting, improves stability and robustness, and allows incorporating prior knowledge or constraints when dealing with noisy or high-dimensional data.

As shown in Table~\ref{tab:abalation_reg}, we varied $\epsilon$ from $e^{-3}$ to 1. When $\epsilon$ was between 1 and 10, the number of mapped distribution data points was less than 100. We also experimented with $\epsilon$ set as $e^{-4}$ and 0 in four groups. With $\epsilon = e^{-4}$, the final optimal mapping data points were all less than 50, and with $\epsilon = 0$, no mapping was obtained in any of the four groups, highlighting the importance of the regularization term. Different values for the regularization term slightly affected the performance of the four groups. \textbf{A2B} and \textbf{Q2A} achieved the best performance with $\epsilon = e^{-3}$, \textbf{A2Q} performed best with $\epsilon = e^{-2}$, and \textbf{Q2B} achieved its best performance with $\epsilon = e^{-1}$.

\begin{table}
	\centering
	\caption{\small Ablation studies on regularization term $\epsilon$. Each row represents the MAE and RMSE for different cross-domain settings when $\epsilon$ is set to $e^{-3}$, $e^{-2}$, $e^{-1}$, and 1, respectively.}
	\small
	\setlength{\tabcolsep}{2.5pt}
	\begin{tabular}{lcccccccc}
	\toprule
	\multirow{2}[2]{*}{Reg-term} & \multicolumn{2}{c}{A2B} & \multicolumn{2}{c}{A2Q} & \multicolumn{2}{c}{Q2A} & \multicolumn{2}{c}{Q2B}\\
	\cmidrule(lr){2-3} \cmidrule(lr){4-5}\cmidrule(lr){6-7}\cmidrule(lr){8-9}
	 & MAE & RMSE & MAE & RMSE & MAE & RMSE & MAE & RMSE\\
	\midrule
	$\epsilon = e^{-3}$ & \textbf{10.9} & 18.8 & 118.5 & \textbf{214.9} & \textbf{67.0} & \textbf{128.4} & 12.4 & 21.4\\
	$\epsilon = e^{-2}$ & 11.5 & 20.3 & \textbf{113.9} & 215.6 & 72.4 & 136.2 & 12.8 & 22.9\\
	$\epsilon = e^{-1}$ & 11.9 & 20.2 & 116.3 & 217.3 & 73.7 & 142.1 & \textbf{11.3} & \textbf{19.6}\\
	$\epsilon = 1$ & 11.1 & \textbf{18.4} & 121.1 & 219.7 & 72.1 & 139.1 & 13.0 & 23.1\\
	\bottomrule
	\end{tabular}
\label{tab:abalation_reg}
\end{table}%
\begin{figure}
	\centering
	\includegraphics[width = 0.89\columnwidth]{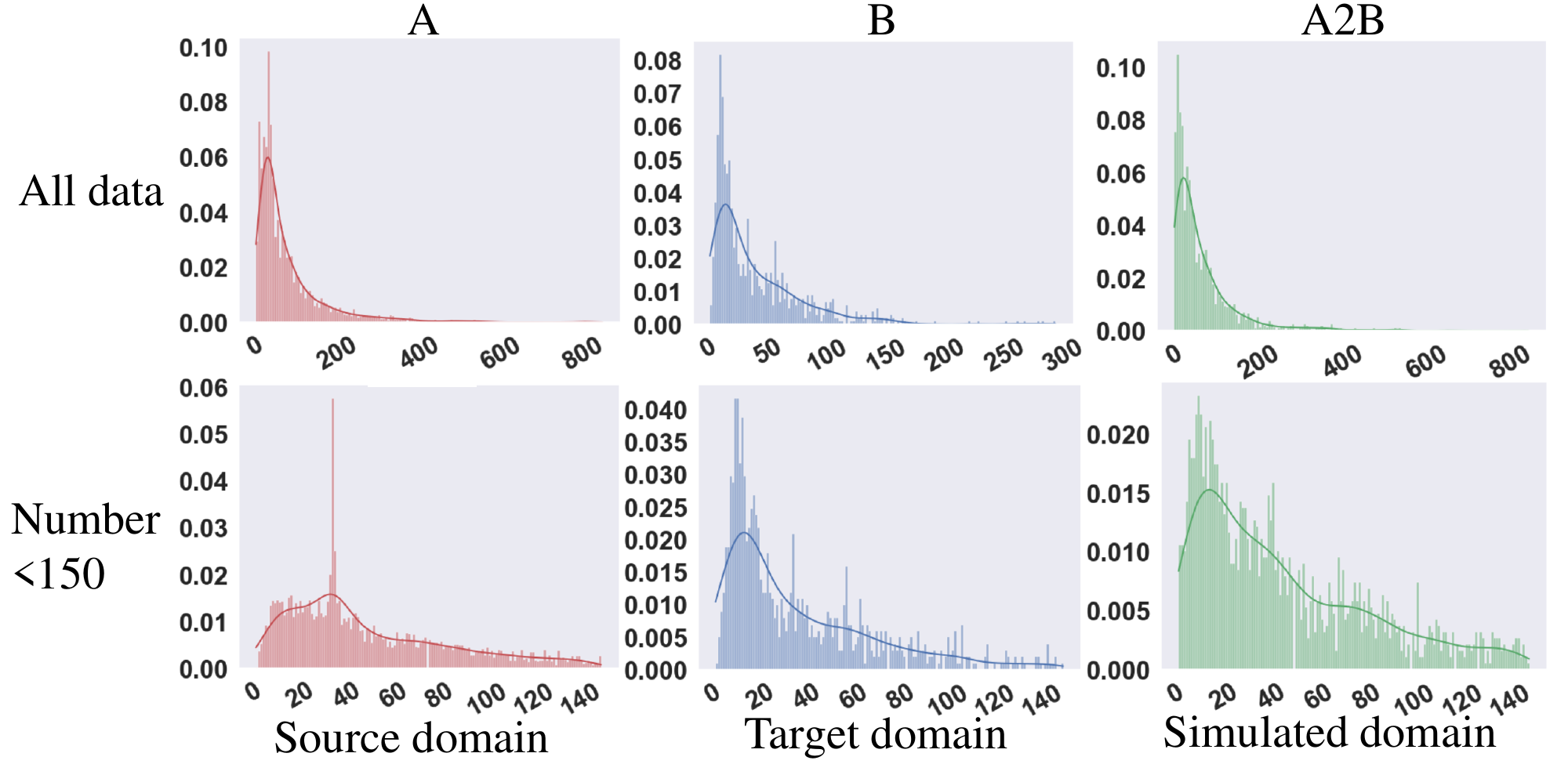}
	\caption{\small Visualization of frequency histograms depicting the number of people in the source domain distribution, target domain distribution, and simulated distribution (A2B).}
	\label{fig:fre2}
\end{figure}

\section{Analysis of Domain Alignment}
To demonstrate the density frequency before and after the transfer, we visualize frequency distribution histograms of the source and target domain distributions and the simulated distribution histograms after transformation, as shown in Fig.~\ref{fig:fre2}. Since our distribution transfer is at the regional level, we show distribution transfer when the region is set to 256. Before distribution transfer, the overall distribution of datasets \textbf{A} and \textbf{B} is concentrated in range of 0 to 150, so we show frequency distribution histograms before and after the transfer for images with 0 to 150 people. From the figure, it can be seen that \textbf{A} has a very high peak around 40, while the distribution of \textbf{B} peaks at around 10 and exhibits a sudden increase in a few specific regions. After distribution transfer, the distribution is closer to the distribution of \textbf{B}, peaking at around 10 and trending similarly to \textbf{B} overall.
\end{document}